\documentclass[10pt,twocolumn,letterpaper]{article}

\usepackage[pagenumbers]{cvpr} 

\definecolor{cvprblue}{rgb}{0.21,0.49,0.74}
\usepackage[pagebackref,breaklinks,colorlinks,allcolors=cvprblue]{hyperref}

\def\paperID{5798} 
\def\confName{CVPR}
\def\confYear{2025}

\title{Prompt-Guided Mask Proposal for Two-Stage Open-Vocabulary Segmentation}

\author{
Yu-Jhe Li$^{1}$\thanks{~Work done during 2023 summer internship at Adobe.} \quad
Xinyang Zhang$^{2*}$
\quad Kun Wan$^{1*}$
\quad Lantao Yu$^{1*}$
\quad Ajinkya Kale$^{1*}$
\quad Xin Lu$^{3*}$
\\
$^{1}$Adobe
\qquad\qquad
$^{2}$Amazon
\qquad\qquad
$^{3}$ByteDance\\
{\tt\small $^{*}$Work done in summmer 2023 during Yu-Jhe Li's internship with Adobe}
}
\begin{document}
\twocolumn[{%
\renewcommand\twocolumn[1][]{#1}%
\maketitle
\vspace{-10mm}
\begin{center}
    \centering
    \includegraphics[width=0.95\textwidth]{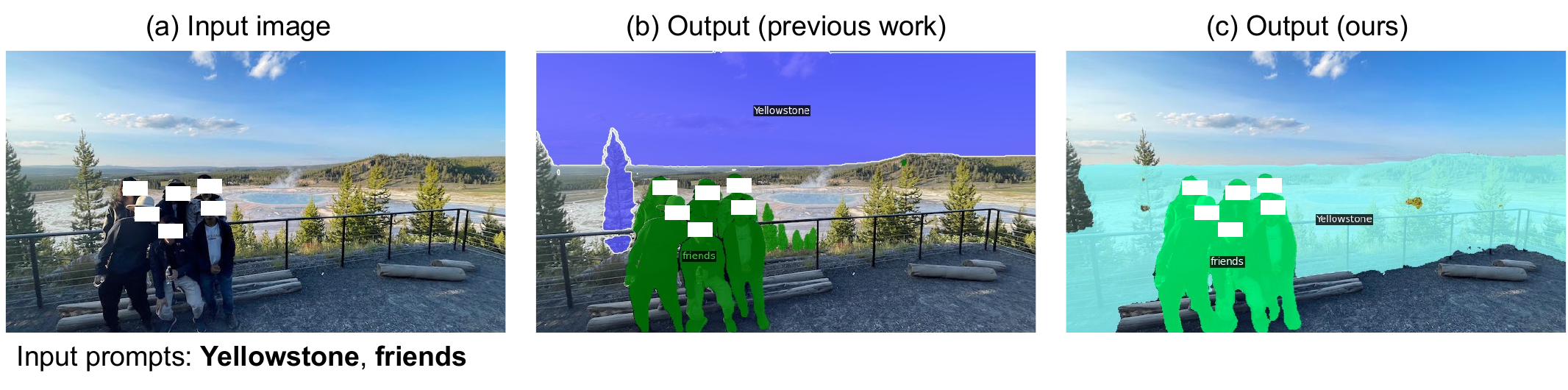}
    \vspace{-4mm}
    \captionof{figure}{The significance of prompt-guided mask proposals for open-vocabulary segmentation. Compared with the previous work (\cite{liang2023open} as an example in the middle image), our proposed mask proposals with input prompt guidance contain the reasonable segmentation mask, which allows the CLIP model to retrieve the proprietary prompt such as ``Yellowstone''. Faces are masked out for the privacy reason.}
    \label{fig:show}
\end{center}%
}]

\begin{abstract}

We tackle the challenge of open-vocabulary segmentation, where we need to identify objects from a wide range of categories in different environments, using text prompts as our input. To overcome this challenge, existing methods often use multi-modal models like CLIP, which combine image and text features in a shared embedding space to bridge the gap between limited and extensive vocabulary recognition, resulting in a two-stage approach: In the first stage, a mask generator takes an input image to generate mask proposals, and the in the second stage the target mask is picked based on the query. 
However, the expected target mask may not exist in the generated mask proposals, which leads to an unexpected output mask. In our work, we propose a novel approach named Prompt-guided Mask Proposal (PMP) where the mask generator takes the input text prompts and generates masks guided by these prompts. Compared with mask proposals generated without input prompts, masks generated by PMP are better aligned with the input prompts.
To realize PMP, we designed a cross-attention mechanism between text tokens and query tokens which is capable of generating prompt-guided mask proposals after each decoding. We combined our PMP with several existing works employing a query-based segmentation backbone and the experiments on five benchmark datasets demonstrate the effectiveness of this approach, showcasing significant improvements over the current two-stage models ($1\% \sim 3 \%$ absolute performance gain in terms of mIOU). The steady improvement in performance across these benchmarks indicates the effective generalization of our proposed lightweight prompt-aware method.  

\end{abstract}

\section{Introduction}

We are addressing the challenge of open-vocabulary segmentation, aiming to segment specified objects based on input text prompts. 
Open-vocabulary segmentation~\cite{li2022languagedriven,xu2021simple,ghiasi2022scaling,ding2023open} was proposed to overcome the constraints of closed-vocabulary segmentation that predicts a set of non-overlapping masks labeled with a limited number of classes. These approaches use text embeddings of category names~\cite{zareian2021open}, represented in natural language, as label embeddings, instead of learning them from the training dataset. This allows models to identify objects from a broader vocabulary, thus improving their ability to generalize to unseen categories. To ensure meaningful embeddings, a pretrained text encoder~\cite{radford2021learning,raffel2020exploring,liu2019roberta,devlin2018bert} is typically employed, effectively capturing the semantic meaning of words and phrases, which is critical for open-vocabulary segmentation.

Recently, several studies propose utilizing pre-trained vision-language models, such as CLIP~\cite{radford2021learning}, for open-vocabulary segmentation~\cite{li2022languagedriven,xu2021simple,ghiasi2022scaling,ding2023open}. Particularly, two-stage methods have demonstrated significant promise: initially generating class-agnostic mask proposals and subsequently employing pre-trained CLIP for open-vocabulary classification. The effectiveness of these approaches relies on two assumptions: (1) the model's ability to generate class-agnostic mask proposals and (2) the transferability of pre-trained CLIP's classification performance to masked image proposals. Recent methods like SimBaseline~\cite{xu2021simple}, OVSeg~\cite{liang2023open}, and 
ODISE~\cite{xu2023open} adopt a two-stage framework to adapt CLIP for open-vocabulary segmentation. In these methods, images undergo initial processing by a robust mask generator, such as Mask2Former~\cite{cheng2022masked} or MaskRCNN~\cite{he2017mask}, to obtain mask proposals. Subsequently, each masked image crop or embedding is generated and input into a frozen CLIP model for classification. However, these models often assume that the generated candidate masks in the first stage consistently contain the correct mask to be retrieved, which is not the case for arbitrary text prompts, as illustrated in Figure~\ref{fig:show} (b). For example, if the text prompt is the subject word ``Yellowstone'', the model may not be able to retrieve the region of Yellowstone since the mask proposals used in the original Mask2Former are class-agnostic and most focused on the object-wise region. To produce the ideal result such as Figure~\ref{fig:show} (c), we have to integrate the text-specific information inside the mask proposal to generate the region of the specified mask.

To address the aforementioned issues, we propose a novel approach \textbf{Prompt-guided Mask Proposal (PMP)} where the mask generator takes the input text prompts into account and generates masks guided by these prompts for existing two-stage models. Specifically, we integrate text tokens from the input prompts alongside query tokens in the end-to-end transformer-based mask generators (\textit{i.e.}, Mask2Former~\cite{cheng2022masked} and MaskFormer~\cite{cheng2021per}). Besides each of the standard cross-attention decoding in the transformer decoder, we propose our designed cross-attention mechanism between text tokens and query tokens, and the new query tokens are used to generate mask embeddings for mask proposals after each decoding process. We believe this mechanism will allow the query tokens to take the prompt-specific information into account and is able to generate the prompt-specific mask proposals for the second stage. Hence, our proposed PMP is capable of recognizing the masked region with arbitrary text prompts instead of limited class names as existing works. We combine our PMP with the several existing works employing query-based mask proposals and the experiments on five benchmark datasets demonstrate the effectiveness of this approach, showcasing significant improvements over the current state-of-the-art models. This also highlights the generalization of our proposed prompt-aware pipeline. The contributions of this paper are summarized below:

\begin{itemize}
    \item We have unveiled the issue in the class-agnostic mask proposals in the existing models of two-stage open vocabulary segmentation.
    \item We propose a prompt-guided mask proposal on top of the current end-to-end mask proposal framework, which produces prompt-specific proposals for mask classification in the second stage.
    \item Our model serves as a lightweight prompt-aware adaptor that boosts existing open-vocabulary segmentation models on the alignment between the output mask with the input prompt. 
    The performance gains in the experiments with multiple models of prior art support the effectiveness of our proposed method.
\end{itemize}
\section{Related Works}

\paragraph{\textbf{Vision-Language Pre-trained Model.}}

Vision-language models aim to encode both vision and language in a unified model. Initial approaches~\cite{tan2019lxmert,zhang2021vinvl,chen2020uniter} involve extracting visual representations using pre-trained object detectors, fine-tuning them on downstream tasks with language supervision. Recent advancements in this domain, spurred by large language models like BERT~\cite{devlin2018bert} and GPT~\cite{brown2020language}, have shown that pretraining dual-encoder models on large-scale noisy image-text pairs with contrastive objectives, as demonstrated by CLIP~\cite{radford2021learning} and ALIGN~\cite{jia2021scaling}, can yield representations with strong cross-modal alignment. Subsequent works~\cite{yuan2021florence,yu2022coca,alayrac2022flamingo} further validate these findings, achieving impressive results in zero-shot transfer learning, such as open-vocabulary image recognition.

\paragraph{\textbf{Segmentation.}}
Segmentation can be categorized into semantic, instance, and panoptic segmentation based on the semantics of grouping pixels. Semantic segmentation interprets high-level category semantic concepts, treating the task as a per-pixel classification problem~\cite{chen2014semantic,ronneberger2015u,chen2017deeplab,chen2017rethinking,chen2018encoder,fu2019dual,gu2022multi,xie2021segformer,yuan2020object,zheng2021rethinking}. Instance segmentation involves grouping foreground pixels into different object instances, often addressing the task with mask classification~\cite{kirillov2017instancecut,liu2018path,cai2018cascade,bolya2019yolact,chen2019hybrid,tian2020conditional,wang2020solov2,qiao2021detectors}. Panoptic segmentation seeks holistic scene understanding, decomposing the problem into various proxy tasks and merging the results~\cite{kirillov2019panoptic,liu2019end,kirillov2019panoptic,xiong2019upsnet,cheng2020panoptic,li2020unifying,wang2020axial,chen2020scaling}. Recent works, following the end-to-end approach of DETR~\cite{carion2020end}, \cite{wang2021max,strudel2021segmenter,cheng2021per,cheng2022masked,li2022panoptic,yu2022cmt,yu2022k,jain2023oneformer,li2023mask} build on the idea of mask classification using pixel and mask decoders, as in Mask2Former\cite{cheng2022masked}, and integrate text tokens into the decoding process. Similarly, our proposed method builds on top of the pixel decoder and mask decoder of Mask2Former~\cite{cheng2022masked} by exploiting the text tokens in the decoding process. 

\begin{figure*}[t!]
  \centering
  \includegraphics[width=\linewidth]{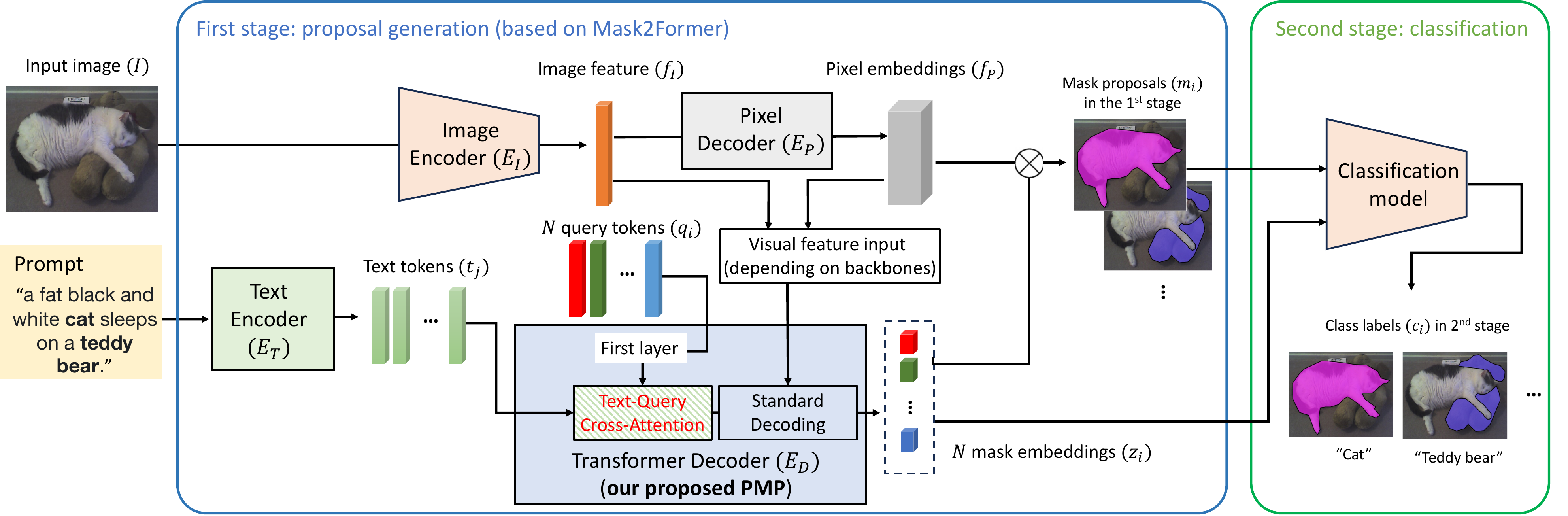}
  \captionsetup{aboveskip=4pt}\captionsetup{belowskip=0pt}
  \caption{\textbf{Overview of the proposed prompt-guided mask proposal (PMP) in the two-stage pipeline for open vocabulary segmentation}. The entire pipeline contains an image encoder $E_I$, a pixel decoder $E_P$, a text encoder $E_T$, and a transformer decoder $E_D$. We utilize the query-based transformer decoder $E_D$ to produce the $N$ mask embeddings $\{ {z}_i \}_{i=1}^N$ given $N$ queries $\{ {q}_i \}_{i=1}^N$. Specifically, our PMP built on top of the transformer decoder takes the N query tokens and the essential text tokens $\{ {t}_j \}_{j=1}^M$, where $M$ varies depending on the number of given prompts, to produce the $N$ mask embeddings from a given image ($I$). It consists of a stack of layers, each is built with a text-query cross-attention block followed by a standard decoding block. The image encoder $E_I$ is introduced to obtain a visual-spatial feature $f_I$ of the entire image for the transformer encoder to obtain the $N$ mask embeddings from an image. The transformer decoder is also able to take multi-level pixel embeddings $f_P$ generated by the introduced pixel decoder $E_P$ for improved generalization. The generated mask embeddings can be transformed into mask proposals $\{ \Tilde{m}_i \}_{i=1}^{N}$ by the multiplication with the pixel embeddings $f_P$. The class labels $\{ \Tilde{c}_i \}_{i=1}^N$ are also produced by these mask embeddings with the pre-trained language model (\textit{e.g.,} CLIP).}
  \label{fig:model}
  \vspace{-4mm}
\end{figure*}

\paragraph{\textbf{Open-Vocabulary Segmentation.}}

Open-vocabulary segmentation targets segmenting arbitrary classes, including those inaccessible during training. Prior works~\cite{li2022languagedriven,ghiasi2022scaling,xu2021simple,liang2023open,ding2022decoupling,xu2022groupvit,zhou2022extract,xu2023side,zou2023generalized,zhou2023zegclip} achieve open-vocabulary semantic segmentation by leveraging large pre-trained vision-language models~\cite{radford2021learning,rombach2022high,jia2021scaling}. Recent two-stage approaches like MaskCLIP~\cite{ding2023open} introduce a pipeline with a class-agnostic mask generator and a frozen CLIP encoder for cross-modal alignment, expanding CLIP's scope to open-vocabulary panoptic segmentation. ODISE~\cite{xu2023open} leverages the innate potential of pre-trained text-image diffusion models~\cite{rombach2022high} for robust open-vocabulary panoptic segmentation. 
For one-stage approaches, FC-CLIP~\cite{yu2023convolutions} proposes a single-stage framework using a single frozen convolutional CLIP backbone while CAT-seg~\cite{cho2024cat} leverages multi-scale CLIP feature aggregation for pixel-level segmentation. In this paper, we focus on improving the quality of mask proposal in two-stage approaches since these models often assume that the generated candidate masks in their proposals always contain the correct mask to be retrieved, which is not the case for arbitrary text prompts.
\section{The Proposed Method}
\label{sec:method}

\subsection{Overview}


\paragraph{Problem Formulation.} Given an image $I \in \mathbb{R}^{H\times W \times 3}$, the objective of open-vocabulary segmentation is to divide it into a set of $K$ masks, each paired with a semantic label: $\{ (m_i,c_i)\}^K_{i=1}$. Each mask, denoted as $m_i \in \{0,1\}^{H \times W}$, represents a binary indication of the area inside the entire image, and it is associated with a corresponding class label $c_i$. During the training phase, a fixed set of class labels $\mathcal{C}_{train}$ is utilized. However, during the inference phase, a different set of categories $\mathcal{C}_{test}$ is employed. In the open-vocabulary scenario, $\mathcal{C}_{test}$ may include novel categories that were not present during training, i.e., $\mathcal{C}_{train} \neq \mathcal{C}_{test}$. Initially, we adhere to the approach of prior works~\cite{xu2023open,liang2023open,ding2023open}, assuming the pre-selection of category names from $\mathcal{C}_{test}$ is available during testing on benchmarks. Furthermore, it is important to highlight that the paper introduces a more challenging setting with more abstract testing categories and prompts. This setting is more practical for real-world applications.

\paragraph{Overview of Two-Stage Pipeline.} In two-stage open vocabulary segmentation following previous works (MaskCLIP~\cite{ding2023open}, OVSeg\cite{liang2023open}, ODISE~\cite{xu2023open}), we generate the $N$ mask proposals $\{ \Tilde{m}_i \}_{i=1}^N$ in the first stage, where $ \Tilde{m}_i \in \mathbb{R} ^{H \times W} $. We then leverage the pre-trained text-image model (\textit{e.g.,} CLIP~\cite{radford2021learning}) to classify these proposals $\{ \Tilde{m}_i \}_{i=1}^N$ into class labels: $\{ \Tilde{c}_i \}_{i=1}^N$, where $ \Tilde{c}_i \in \mathbb{R} ^{|\mathcal{C}|} $ and $\mathcal{C}$ refers to the selected classes or prompts. Specifically, $\mathcal{C} = \mathcal{C}_{train}$ during the training stage and  $\mathcal{C} = \mathcal{C}_{test}$ during the test stage. We can set $\mathcal{C} = \mathcal{C}_{prompt}$ for random input prompt class for real-world applications. As we mentioned earlier, the quality of the class agnostic masks $\{ \Tilde{m}_i \}_{i=1}^N$ in the first stage will affect the location of the class-specific segmented region. If the correct mask is not within the mask proposals $\{ \Tilde{m}_i \}_{i=1}^N$, the open-vocabulary classification model will not produce the correct result in the second stage anymore. Hence, we propose to improve the quality of the proposals in the first stage and propose a Prompt-guided Mask Proposal (PMP) as shown in Figure~\ref{fig:model}. 

\paragraph{\textbf{Overview of PMP.}} As we present the overview of our PMP module in the two-stage pipeline in Figure~\ref{fig:model}, the entire pipeline also contains an image encoder $E_I$, a pixel decoder $E_P$, a text encoder $E_T$, and a transformer decoder $E_D$. Following MaskFormer~\cite{cheng2021per} and Mask2Former~\cite{cheng2022masked}, we utilize the query-based transformer decoder $E_D$ to produce the $N$ mask embeddings $\{ {z}_i \}_{i=1}^N$ given $N$ queries $\{ {q}_i \}_{i=1}^N$. Specifically, the transformer decoder takes the N query tokens and the essential text tokens $\{ {t}_j \}_{j=1}^M$, where $M$ varies depending on the number of given prompts, to produce the $N$ mask embeddings from a given image ($I$). Inside the Transformer decoder, the key contribution of PMP is our designed text-query cross attention mechanism~(highlighted). The image encoder $E_I$ is introduced to obtain a visual-spatial feature $f_I$ of the entire image for the transformer decoder to obtain the $N$ mask embeddings from an image. The transformer decoder is also able to take multi-level pixel embeddings $f_P$ generated by the introduced pixel decoder $E_P$ (following Mask2Former~\cite{cheng2022masked}) for improved generalization. Therefore, the generated mask embeddings can be transformed into mask proposals $\{ \Tilde{m}_i \}_{i=1}^N$ by the multiplication with the pixel embeddings $f_P$. The class labels $\{ \Tilde{c}_i \}_{i=1}^N$ are also produced by these mask embeddings with the pre-trained language model.

\paragraph{\textbf{Inference.}} Similar to the previous work (ODISE\cite{xu2023open}), the input format can be either a series of class names or a sentence. If the input format is sentences or captions, we extract the nouns from the sentence as the processed class names. Since the trained dataset COCO-stuff provides both class names and captions, our model can take both formats during the training and testing stages, which has the flexibility based on user request. 
Based on our experimental experience on one Nvidia V100, extracting the nouns and the CLIP embeddings for each noun takes 0.2 seconds for 20 tokens and 0.8 seconds for 100 tokens with batch processing, which ends up with $\sim$1s in the entire pipeline.

\subsection{Preliminary of Two-Stage Pipeline}

We now provide the more context of the standard two-stage open-vocabulary segmentation model built on top of previous works. It comprises a segmentation component responsible for generating mask proposals and an open-vocabulary classification model. In alignment with prior research~\cite{liang2023open,xu2023open}, our model builds upon the foundations laid by MaskFormer~\cite{cheng2021per} and Mask2Former~\cite{cheng2022masked}. Diverging from conventional per-pixel segmentation methods, MaskFormer~\cite{cheng2021per} produces $N$ mask proposals and corresponding class predictions through $N$ learnable query tokens $\{ {q}_i \}_{i=1}^N$. This pipeline resembles a query-based end-to-end approach akin to the principles of DETR~\cite{carion2020end} in the context of object detection. Each proposal is represented by an $H \times W$ binary mask, denoting the spatial extent of the target object. Initially, the class prediction constitutes a $C$-dimensional distribution, where $C$ signifies the number of classes in the training set.


\noindent\textbf{Classification from Mask Embedding.} To tailor the backbone for the open-vocabulary scenario, as outlined in~\cite{liang2023open,xu2023open}, this backbone undergoes modifications so that it generates a $N_c$-dimensional proposal embedding for each mask. Here, $N_c$ corresponds to the embedding dimension of a pre-trained text-image model (e.g., $512$ for ViT-B/16 and $768$ for ViT-L/14 in CLIP). This adjustment enables MaskFormer to undertake open-vocabulary segmentation. In particular, if we intend to categorize the mask into $K$ classes, we can employ a CLIP model's text encoder to produce $K$ text embeddings for each class, denoted as $\{t_k | t_k \in \mathbb{R}^{N_c} \}_{k=1}^K$. Subsequently, we assess each mask embedding $z_i$ against the text embeddings and predict the probability of the $k^{th}$ class using the softmax function:
\begin{equation}
    p_{i,k} = \frac{\exp (\sigma(z_i,t_k)/\tau)}{\sum_k \exp(\sigma(z_i,t_k)/\tau)},
\end{equation}
where $\sigma(\cdot,\cdot)$ represents the cosine similarity between two embedding vectors, and $\tau$ is the temperature coefficient~\cite{radford2021learning}. For instance, when training the modified query-based pipeline on the COCO-Stuff dataset~\cite{caesar2018coco}, we would have $K=171$ classes and $171$ CLIP text embeddings. Additionally, we would append a $172^{nd}$ learnable embedding $\phi$ to signify the category of ``no object" or ``background."


\noindent\textbf{Classification from Visual Embeddings.} Moreover, in line with the approaches proposed in ODISE~\cite{xu2023open} and OVSeg~\cite{liang2023open}, the efficacy of the classification in the second stage can be further heightened by integrating it once again with a text-image discriminative model, such as CLIP~\cite{radford2021learning}. Consequently, we also utilize a text-image discriminative model, specifically CLIP image encoder $E_I^{CLIP}$, to perform additional classification on each predicted masked region of the original input image into one of the test categories.
To elaborate, given an input image $I$, we initially encode it into a feature map using the image encoder $E_I^{CLIP}$ of a text-image discriminative model. Subsequently, for a mask $m_i$ predicted by the two-stage model for image $I$, we aggregate all the features at the output of the image encoder $E_I^{CLIP}$ that fall within the predicted mask $m_i$ to compute a mask-pooled image feature $v_i$. Each mask-pooled feature $v_i$ is then compared with the text embedding, and the probability of the $k^{th}$ class is predicted using the softmax function:
\begin{equation}\label{eq:2}
    \hat{p}_{i,k} = \frac{\exp(\sigma(v_i,t_k)/\tau)}{\sum_k \exp(\sigma(v_i,t_k)/\tau)}.
\end{equation}
The geometric mean of the category predictions from the second stage and discriminative models can be defined as:
\begin{equation}\label{eq:3}
{p}_{i,k}^{out} = {p}_{i,k}^{(\lambda)}* \hat{p}_{i,k}^{(1-\lambda)},
\end{equation}
where $\lambda \in [0,1]$.

\subsection{Prompt-guided Proposal (PMP) Generation}

As mentioned earlier, the mask proposal generator trained in the two-stage end-to-end pipeline is unable to produce class-specific masks because the definition of an object is constrained by the class definitions in the training set. For instance, if the training set only encompasses the class ``vehicle", it is unlikely that the model will automatically segment a vehicle into finer parts such as ``tire", ``windshield", or ``light" due to the absence of class-specific mask proposals. This leads to the challenge of missing proposals in the second stage. Consequently, devising a strategy to train a zero-shot model capable of generating class-specific mask proposals poses a significant challenge.

In order to avoid the missing ideal mask proposals in the second stage, we propose to improve the quality of the mask proposals in the first stage and present our pipeline in Figure~\ref{fig:model}. Our prompt-guided mask proposal is built on top of the end-to-end query-based segmentation models (MaskFormer~\cite{cheng2021per} or Mask2Former~\cite{cheng2022masked}). 
Similar to those methods, our first stage also has a backbone (image encoder), a pixel decoder, and a Transformer decoder. 

\subsubsection{Text-Query Cross Attention.}
The innovation to make the query-based pipeline prompt-specific or class-specific is we introduce the text tokens ($t_j$) for the masked attention in the transformer decoder. Specifically, the standard cross-attention with the residual path in the transformer encoder can be originally defined as:

\begin{equation}\label{eq:cross}
    {X}_l = \mathrm{softmax} ({Q}_l {K}_l^T) {V}_l + {X}_{l-1},
\end{equation}
where $l$ indicates the layer index and $X_l \in \mathbb{R}^{N \times C}$ indicates the $N$ $C$-dimensional query features at the $l^{th}$ layer. In addition, ${Q}_l = f_Q ({X}_l) \in \mathbb{R}^{N \times C}$ is the transformed query features from the query features $X_l$, while ${K}_l = f_K ({f}_I) \in \mathbb{R}^{H_l W_l \times C}$ and ${V}_l = f_V ({f}_I) \in \mathbb{R}^{H_l W_l \times C}$ are the transformed features from the image features. The $X_0$ is set to be the query tokens $\{ {q}_i \}_{i=1}^N$ in the beginning. $f_Q (\cdot)$, $f_K (\cdot)$, and $f_V (\cdot)$ are the linear transformation functions.


In order to ensure our improved version of the transformer decoder is able to produce the mask embeddings $\{ {z}_i \}_{i=1}^N$ that are conditioned on the input text tokens, we apply another cross-attention between the text tokens $\{ {t}_j\}_{M=1}^N$ and the query feature before the standard cross-attention step in the transformer encoder. That is:
\begin{equation}
\begin{split}
    {Q}_l' = &~ \mathrm{softmax} ({Q}_l {K}_t^{\intercal}) {V}_t\\
    {X}_l = &~ \mathrm{softmax} ({Q}_l' {K}_l^{\intercal}) {V}_l + {X}_{l-1},
\end{split}
\end{equation}
where ${K}_t = f_K^t ( \{ {t}_j\}_{M=1}^N) \in \mathbb{R}^{M \times C}$ and ${V}_t = f_V^t (\{ {t}_j\}_{M=1}^N) \in \mathbb{R}^{M \times C}$ are the transformed features from the text tokens $\{ {t}_j\}_{M=1}^N$. $f_K^t (\cdot)$ and $f_V^t (\cdot)$ are the linear transformation functions. Note that this revised version of cross-attention can be built on top of Mask2Former~\cite{cheng2022masked} without further efforts by applying a similar cross-attention before the masked attention in the transformer decoder and having the image features ${f_I}$ replaced with pixel features ${f_P}$ from pixel decoder. Note that positional embeddings and predictions from intermediate Transformer decoder layers are omitted here for readability. 

We would like to note that, our proposed method is simple yet effective for producing prompt-specific mask proposals for the second stage. Since our proposed method can be built on top of the Mask2Former~\cite{cheng2022masked}, our methods can definitely serve as a lightweight prompt-aware component for most of the state-of-the-art approaches that employ Mask2Former. Later in the experiments, we will present the results and comparisons using our proposed prompt-guided proposals.

\section{Experiment}
\label{sec:EXP}

\begin{table*}[t!]
  
  \centering
  \captionsetup{aboveskip=0pt}\captionsetup{belowskip=0pt}
  \caption{Comparison with state-of-the-art two-stage methods in open-vocabulary settings on five benchmark datasets. The mIOU (\%) is utilized as an evaluation protocol for each of the five benchmarks. The number in bold indicates the best results.}\label{table:ovseg}
  \resizebox{0.8\linewidth}{!}
  {
  \begin{tabular}{l|ccccc}
  \toprule
  %
  Method & ADE-847 & PC-459 & ADE-150 & PC-59 & VOC\\
  \midrule
  FC-CLIP~\cite{yu2023convolutions} (ConvNeXt-Large) & 14.8 & 18.2 & 34.1 & 58.4 &  95.4\\
  FC-CLIP~\cite{yu2023convolutions} (ConvNeXt-Large) + PMP (Ours) & \textbf{16.2} {\color{blue} (+1.4)} & \textbf{19.1} {\color{blue} (+0.9)} & \textbf{35.3} {\color{blue} (+1.2)} & {59.9} {\color{blue} (+1.5)} & \textbf{96.0} {\color{blue} (+0.6)}\\
  FC-CLIP~\cite{yu2023convolutions} (ResNet50x64) & 10.8 & 16.2 & 28.4 & 55.7 & 95.1 \\
  FC-CLIP~\cite{yu2023convolutions} (ResNet50x64) + PMP (Ours) & 12.9 {\color{blue} (+2.1)} & 17.5 {\color{blue} (+1.3)} & 30.5 {\color{blue} (+2.1)} & 56.9 {\color{blue} (+1.2)} & 95.8 {\color{blue} (+0.7)}\\
  \midrule
  SAN~\cite{xu2023side} & 12.4 & 15.7 & 32.1 & 57.7 & 94.6\\
  SAN~\cite{xu2023side} + PMP (Ours) & 13.1 {\color{blue} (+0.7)} & 16.8 {\color{blue} (+1.1)} & 33.5 {\color{blue} (+1.4)} & 60.1 {\color{blue} (+2.4)} & 95.1 {\color{blue} (+0.5)}\\
  SAN~\cite{xu2023side} w/o query pos & 12.8 & 15.3  & 32.5  & 57.1  & 94.2 \\
  SAN~\cite{xu2023side} + PMP (Ours) w/o query pos & 14.2 {\color{blue} (+1.4)} & 17.9 {\color{blue} (+2.6)} & 34.1 {\color{blue} (+1.6)} & \textbf{61.0} {\color{blue} (+2.9)} & 95.5 {\color{blue} (+1.3)}\\
  \midrule
  ODISE~\cite{xu2023open} & 11.1 & 14.5 & 29.9 & 57.3 & -\\
  ODISE~\cite{xu2023open} + PMP (Ours) & 11.9 {\color{blue} (+0.8)} & 15.1 {\color{blue} (+0.6)} & 30.8 {\color{blue} (+0.9)} & 58.0 {\color{blue} (+0.7)} & -\\ 
  \midrule
  OVSeg~\cite{liang2023open} & 9.0 & 12.4 & 29.6 & 55.7 & 94.5\\
  OVSeg~\cite{liang2023open} + PMP (Ours) & 12.6 {\color{blue} (+3.6)} & 14.7 {\color{blue} (+2.3)} & 33.5 {\color{blue} (+3.9)} & 57.3 {\color{blue} (+1.6)} & 95.8 {\color{blue} (+1.3)} \\
  
  \bottomrule
  \end{tabular}
  }
  
  
  \vspace{-4.5mm}
\end{table*}

\subsection{Datasets and evaluation protocols}

We perform experiments on six datasets: COCO Stuff~\cite{caesar2018coco}, ADE20K-150~\cite{zhou2017scene}, ADE20K-847~\cite{zhou2017scene}, Pascal Context-59~\cite{mottaghi2014role}, Pascal Context-459~\cite{mottaghi2014role}, and Pascal VOC~\cite{everingham2015pascal}. Following the established practice in previous works~\cite{liang2023open,xu2022simple}, all models undergo training on the COCO Stuff training set and are subsequently evaluated on the remaining datasets. More statistics of these benchmark datasets can be referred to previous works~\cite{li2022languagedriven,ghiasi2022scaling,xu2021simple,liang2023open,ding2022decoupling,xu2022groupvit,zhou2022extract,xu2023side,zou2023generalized,zhou2023zegclip}.

\noindent\textbf{COCO Stuff.} This dataset comprises 164k images with 171 annotated classes, distributed across training (118k images), validation (5k images), and test (41k images) sets. In our experiments, we default to using the entire 118k images from the training set.

\noindent\textbf{ADE20K-150 (ADE-150).}
This is a large-scale scene understanding dataset with 20k training images, 2k validation images, and a total of 150 annotated classes.

\noindent\textbf{ADE20K-847 (ADE-847).} It shares the same images as ADE20K-150 but features a more extensive set of annotated classes (847 classes), presenting a challenging dataset for open-vocabulary semantic segmentation.

\noindent\textbf{Pascal VOC (VOC).} VOC consists of 20 classes of semantic segmentation annotations, with the training set and validation set containing 1464 and 1449 images, respectively.

\noindent\textbf{Pascal Context-59 (PC-59).}
This dataset, designed for semantic understanding, includes 5K training images, 5K validation images, and a total of 59 annotated classes.

\noindent\textbf{Pascal Context-459 (PC-459).}
It shares the same images as Pascal Context-59 but encompasses a more extensive set of annotated classes (459 classes), making it a widely used dataset for open-vocabulary semantic segmentation.

\paragraph{\textbf{Evaluation Protocol.}} Consistent with established practices~\cite{cheng2022masked,xu2022simple,ghiasi2022scaling}, we utilize the mean of class-wise intersection over union (mIOU) in percentage as the metric to evaluate the performance of our models. The pipeline is trained on the COCO Stuff dataset and assessed on the other five datasets for benchmarking results.

\subsection{Implementation of PMP usage}\label{sec:imple}


Since our proposed prompt-guided mask proposals (PMP) are built on top of previous works, we now provide more details for how to combine our PMP with some of the existing two-stage models: FC-CLIP~\cite{yu2023convolutions}, SAN~\cite{xu2023side}, ODISE~\cite{xu2023open}, and OVSeg~\cite{liang2023open}.

\noindent\textbf{OVSeg + PMP.} OVSeg~\cite{liang2023open} is the standard model which utilize Mask2Former as their first stage proposal generation and CLIP as their second stage classification. Hence, to combine PMP with ODISE, we replace the decoding module in their Mask2Former with our proposed PMP decoding module. We follow the same training and inference strategy as their open source code.

\noindent\textbf{ODISE + PMP.} ODISE~\cite{xu2023open} also utilize Mask2Former as their first stage proposal generation and CLIP as their second stage classification. The only difference is ODISE replace the visual feature extraction with stable diffusion. Hence, to combine PMP with ODISE, we also replace the decoding module in their Mask2Former with our proposed PMP decoding module in terms of the implementation in their pipeline.

\noindent\textbf{SAN + PMP.} SAN~\cite{xu2023side} also utilizes several transformer layers to produce the mask proposals and the classification for each proposal. The most difference between SAN's mask decoding model and Mask2Former is that it utilizes CLIP as the vision encoder and feeds CLIP visual feature in each of the transformer layers. To generate  proposals with SAN, the learnable query tokens and visual tokens are first projected as 256-dimension and then used to produce the mask proposals by inner product with visual features from transformer layers. In order to combine our PMP with SAN, we perform additional PMP text-guided cross attention with the learnable queries before its inner product with the visual features. We follow the same training and inference strategy provided by their open-source model.

\noindent\textbf{FC-CLIP + PMP.} FC-CLIP~\cite{yu2023convolutions} also utilizes Mask2Former~\cite{cheng2022masked} as their mask generator, where nine mask decoders are employed to generate the class-agnostic masks by taking as inputs the enhanced pixel features and a set of object queries. We replace their mask decoder with our proposed PMP decoder in their Mask2Former backbone in stage one. For in-vocabulary classification in FC-CLIP in stage two, class embeddings are generated by applying mask-pooling to the pixel features derived from the final output of the pixel decoder, which will be used for produce the classification result with their geometric ensambling. 
For either the training and inference strategy, we follow exactly the same procedure with their open-source code.

\begin{figure*}[t!]
  \centering
  \includegraphics[width=0.9\linewidth]{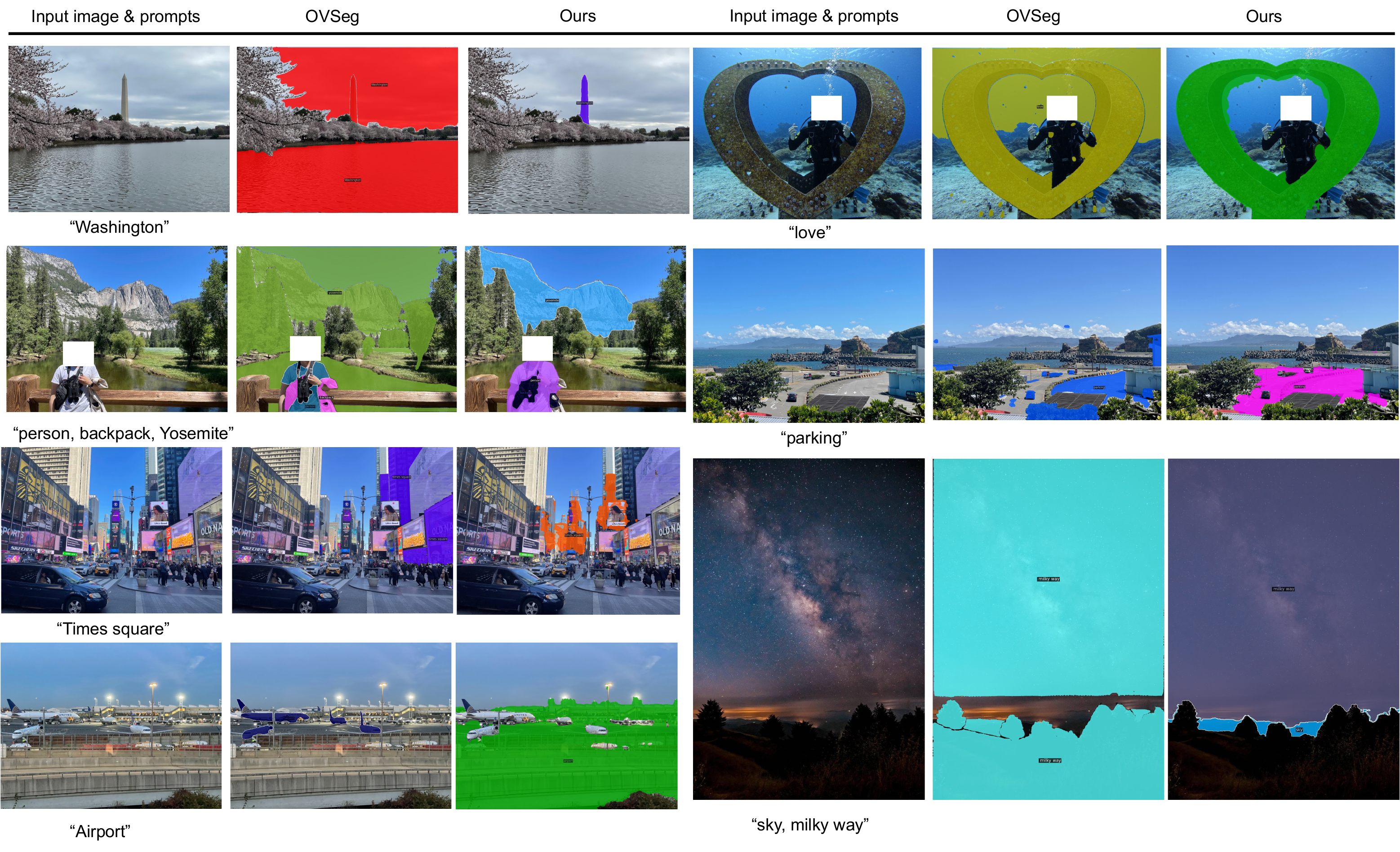}
  \captionsetup{aboveskip=0pt}\captionsetup{belowskip=0pt}
  \caption{Qualitative results of open-vocabulary segmentation on our taken seven example real images. The input prompts contain more than just the object class such as abstract word or proprietary word. We compare with the previous approach OVSeg~\cite{liang2023open}. We'll present more results in the supplementary. Faces are masked out for the privacy reason.
  }
  \label{fig:qual}
  \vspace{-5mm}
\end{figure*}


\subsection{Results and comparison}

\paragraph{\textbf{Quantitative results.}} As we stated earlier, our pipeline serves as a simple yet effective adaptor for generating prompt-specific mask proposals for the existing works. We then compare our proposed method with four current state-of-the-art open vocabulary segmentation methods including FC-CLIP~\cite{yu2023convolutions}, SAN~\cite{xu2023side}, ODISE~\cite{xu2023open}, and OVSeg~\cite{liang2023open} on the five benchmarks: ADE-847, PC-459, ADE-150, PC-59, and PASCAL VOC. Since each of the current four approaches utilizes a different feature encoder training setting, we follow the same architecture and built our prompt-guided proposal on top of each model as mentioned in Sec.~\ref{sec:imple}. We present the result of semantic segmentation in Table~\ref{table:ovseg} and observe some of the phenomena as below. First, compared with all of the existing approaches using MaskFormer (OVSeg) or Mask2Former (FC-CLIP, SAN, ODISE), our improved version of mask proposals bring obvious performance gain (around $1\% \sim 3\%$) among all of them. Second, our method achieves improved performance on top of FC-CLIP with ConvNeXt-Large~\cite{liu2022convnet} backbone. Third, our revised version of SAN with the position embeddings removed in the queries improves the performance.


\begin{figure}[t!]
  \centering
  \includegraphics[width=\linewidth]{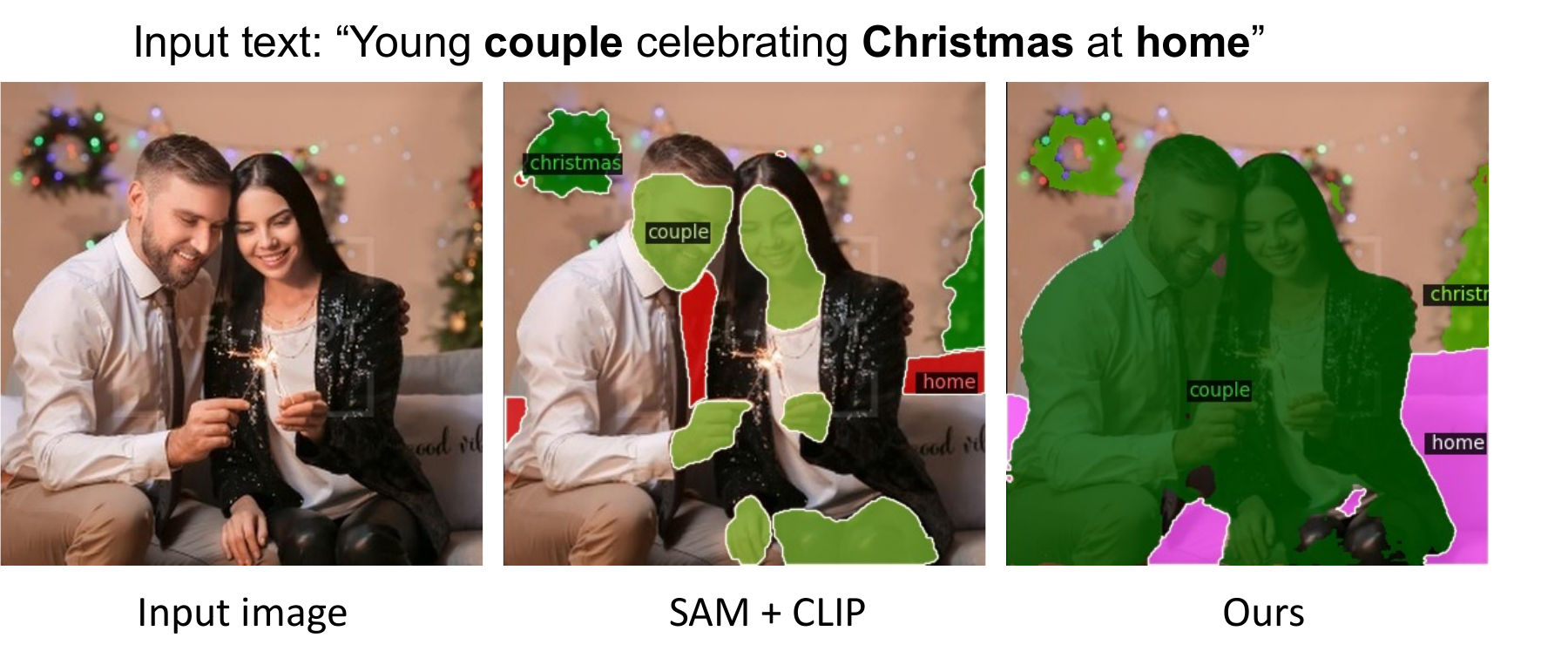}
  \captionsetup{aboveskip=0pt}\captionsetup{belowskip=0pt}
  \caption{Comparison of our model with SAM~\cite{kirillov2023segment} (+CLIP).}
  \label{fig:sam}
  \vspace{-4mm}
\end{figure}

It is worth noting that, the existing five benchmarks only contain the limited object classes that do not appear in the training classes. These benchmarks can not evaluate the performance of the model on some of the general nouns, adjectives, or more abstract words. As shown in Figure~\ref{fig:show}, the word ``Yellowstone'' can not be recognized by each of the existing models without using the prompt-guided proposals. Therefore, even though our model only demonstrates limited performance gain on these benchmarks, it is very effective to generalize the true open-vocabulary segmentation with random text prompts of interest.

Following ODISE~\cite{xu2023open} and FC-CLIP~\cite{yu2023convolutions}, we also provide the results of panoptic segmentation on ADE20k and COCO evaluation dataset in Table~\ref{table:panoptic}. The panoptic segmentation results are evaluated with the panoptic quality (PQ), Average Precision, and mean intersection-over-union (mIoU). The model is only trained on COCO panoptic dataset and we zero-shot evaluate the model on ADE20K. As we observe in the Table~\ref{table:panoptic}, the proposed PMP also brings performance gain the the panoptic benchmarks on top of these two existing methods.

\begin{table}[t!]
  \centering
  \captionof{table}{Ablations of our proposed PMP on open-vocabulary panoptic segmentation with ADE20K and COCO.}  
  \vspace{-2.0mm}
  \resizebox{\linewidth}{!}
  {
  \begin{tabular}{l|ccc|ccc}
  \toprule
  %
  & \multicolumn{3}{c|}{ADE-20k} & \multicolumn{3}{c}{COCO}\\
  Method & PQ & AP  &  mIOU & PQ  & AP & mIOU\\
  \midrule
  ODISE~\cite{xu2023open} & 22.6 & 14.4 & 29.9 & 55.4 & 46.0 & 65.2\\
  ODISE~\cite{xu2023open} + PMP & 24.5  & 15.1  & 33.6  & 56.0  & 46.7  & 65.8\\
  & {\color{blue} (+1.9)} & {\color{blue} (+0.7)} & {\color{blue} (+0.6)} & {\color{blue} (+0.7)} & {\color{blue} (+3.7)} & {\color{blue} (+0.3)} \\
  \midrule
  FC-CLIP~\cite{yu2023convolutions}  & 26.8 & 16.8 & 34.1 & 54.4 & 44.6 & 63.7 \\
  FC-CLIP~\cite{yu2023convolutions}  + PMP & 27.9 & 17.5 & 35.0 & 55.2 & 45.8 & 64.7\\
  & {\color{blue} (+1.1)} & {\color{blue} (+0.7)} & {\color{blue} (+0.9)} & {\color{blue} (+0.8)} & {\color{blue} (+1.2)} & {\color{blue} (+1.0)} \\
  \bottomrule
  \end{tabular}
  }\label{table:panoptic}
\vspace{-3.0mm}
\end{table}

\begin{figure}[t!]
  \centering
  \includegraphics[width=\linewidth]{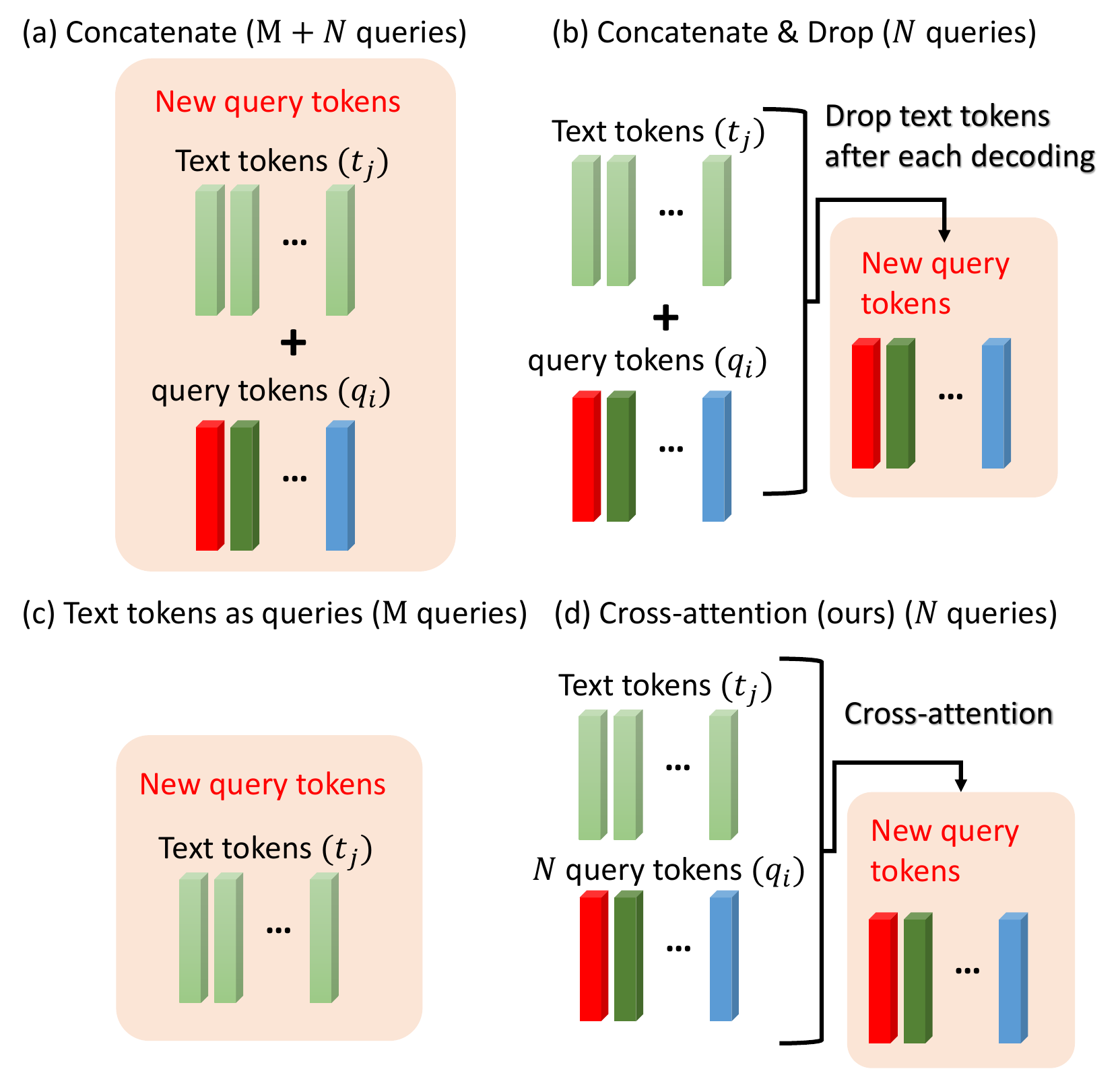}
  \captionsetup{aboveskip=0pt}\captionsetup{belowskip=0pt}
  \caption{Illustration of four different strategies of decoding queries with input text tokens. These strategies include (a) Concatenate, (b) Concatenate and drop, (c) Text tokens as queries, and (d) our proposed cross-attention.}
  \label{fig:abl_decode}
  \vspace{-4mm}
\end{figure}

\begin{table*}[t!]
  
  \centering
  \captionsetup{aboveskip=0pt}\captionsetup{belowskip=0pt}
  \caption{Ablation studies on open-vocabulary settings with different feature strategies of the text tokens as input using OVSeg\cite{liang2023open} as the backbone. The mIOU ($\%$) is utilized as an evaluation protocol for each of the five benchmarks.}\label{table:abl_ovseg}
  \resizebox{0.85\linewidth}{!}
  {
  \begin{tabular}{l|ccccc}
  \toprule
  %
  Method & ADE-847 & PC-459 & ADE-150 & PC-59 & VOC\\
  \midrule
  PMP (Ours) ($N$ queries)& \textbf{12.6} & \textbf{14.7} & \textbf{33.5} & \textbf{57.3} & \textbf{95.8} \\
  PMP (a) concatenate tokens with query ($M + N$ queries) & 11.0 & 13.7 & 30.9 & 56.2 & 94.9\\
  PMP (b) concatenate tokens with query and dropping ($N$ queries) & 9.5 & 12.9 & 30.4 & 56.0 & 95.1\\
  PMP (c) text tokens as queries ($M$ queries) & 10.8 & 13.9 & 31.1 & 56.5 & 95.2\\
  PMP (Ours) w/o text prompts ($N$ queries) & 9.0 & 12.4 & 29.6 & 55.7 & 94.5\\
  \bottomrule
  \end{tabular}
  }
  
  \vspace{-4.5mm}
\end{table*}

\paragraph{\textbf{Qualitative results.}} In order to support the effectiveness of our proposed prompt-guided mask proposal on the true open-vocabulary prompts, we present some of the examples in Figure~\ref{fig:qual}. All of the selected input pictures were real and taken by ourselves during our traveling. As we can observe from the Figure, the model is able to connect the subject ``Washington'' to the Washington Monument in the top-left example. In addition, the model is also able to connect some other subjects such as ``Washington'', ``Yosemite'', ``Times Square'' to the specific scenes inside the images in the remaining examples on the left side. More interestingly, the model is also capable of finding a parking area inside the given image, which indicates further useful cases for real-world application. Lastly, the model is also able to distinguish between ``Milky way'' and ``Sky'' without confusion from the input image regardless of the order of the given prompts. On the other hand, we also present the comparison with SAM~\cite{kirillov2023segment} in Figure~\ref{fig:sam}, which supports the significance of our text-guided mask proposals in the first stage for open-vocabulary segmentation given a caption.

\subsection{Ablation studies}

To further assess our design of a prompt-guided mask proposal, we conducted several ablation studies on different strategies of mask decoding. Besides our proposed cross-attention encoding before each of the cross-attention in the transformer decoder, we also have other candidate decoding strategies to take into the text tokens into our pipeline:
(a) \textbf{Concatenate}, (b) \textbf{Concatenate and drop}, (c) \textbf{Text tokens as queries}.

We present the illustrations for each of the candidate strategies in Figure~\ref{fig:abl_decode}. For the first strategy (a) \textbf{Concatenate}, we directly concatenate the $M$ text tokens with the $N$ learnable query embeddings. This will form $M + N$ query tokens for the transformer encoder to produce the $M + N$ mask embeddings at the end. For the second strategy (b) \textbf{Concatenate \& drop}, similar to (a) we concatenate the text tokens with the learnable query tokens to form $N + M$ tokens before each transformer decoding (cross-attention in Eq.~\ref{eq:cross}) yet we drop the text tokens after the decoding to maintain $N$ query embeddings in each decoding process. Hence, the model still produces $N$ mask embeddings at the end. For the third strategy (c) \textbf{Text tokens as queries}, we solely use the $M$ text tokens as the query tokens in each of the decoding in the transformer decoder. Thus, it will generate $M$ mask embeddings at the end.

We compare these candidate strategies with our designed cross-attention and present the results in Table~\ref{fig:abl_decode}. The results show that our cross-attention is the most optimal strategy among all the candidates for the following reasons. First, even though (a) \textbf{Concatenate} allows the transformer encoder to take into the text tokens, the other $N$ mask embeddings are still not produced conditioned on the $M$ text embeddings. Only the $M$ text tokens are related to the input prompts, which still bring the performance gain compared with no text prompts given (fifth row). Second, (b) \textbf{Concatenate \& drop} seems to take into the text tokens in each of the decoding processes. Yet, the cross-attention in the decoding is mostly calculated individually for each query token, and thus the query embeddings do not benefit from the text tokens that much. For (c) \textbf{Text tokens as queries}, it does help by replacing all the query tokens directly with the text tokens. However, the number of text tokens (\textit{i.e.,} $M < 10$) is much less than the 100 query tokens. Therefore such strategy is not fitted.


\section{Conclusion}
In this work, we have proposed a novel approach named Prompt-guided Mask Proposal (PMP) whose mask generator takes the input text prompts into account and generates masks guided by these prompts for the existing two-stage open-vocabulary segmentation models. The proposed model addressed the issue of the previous assumption that the generated candidate masks may not always contain the target mask for arbitrary text prompts. 
We integrated text tokens with our designed cross-attention mechanism, which achieves optimal text-specific mask production.  The experiments on five benchmark datasets demonstrate the effectiveness of this approach, showcasing significant improvements over the current state-of-the-art models. 
{\small
\bibliographystyle{ieeenat_fullname}
\bibliography{main}
}

\clearpage
\appendix
\section{Appendix}

\subsection{More ablation studies}
In this section we provide more ablation studies on either each stage of pipeline, the backbones, or the hyperparameters to further analyze the sensitivity of them in Table~\ref{table:two-stage} and Table~\ref{table:abl_reb}. We conduct the ablation studies on top of OVSeg~\cite{liang2023open} with our proposed PMP.
Note that the first row if the default setting of the backbone and hyperparameters.

\paragraph{Proposal recall in each stage.}
\begin{table*}[t!]
  \vspace{-8.0mm}
  \centering
  \captionof{table}{Ablation studies on the first and the second stage in open-vocabulary settings on five benchmark datasets. The mIOU (\%) is utilized as an evaluation protocol for each of the five benchmarks.}  
  \vspace{-2.0mm}
  \resizebox{\linewidth}{!}
  {
  \begin{tabular}{l|cc|cc|cc|cc|cc}
  \toprule
  %
  & \multicolumn{2}{c|}{ADE-847} & \multicolumn{2}{c|}{PC-459} & \multicolumn{2}{c|}{ADE-150} & \multicolumn{2}{c|}{PC-59} & \multicolumn{2}{c}{VOC} \\
  Method & first (recall) & second  &  first  (recall) & second  &  first  (recall) &second &  first  (recall) &second  &  first  (recall) & second \\
  \midrule
  OVSeg~\cite{liang2023open} & 17.1 & 9.0 & 25.5 & 12.4 & 41.0 & 29.6 & 66.2 & 55.7 & 95.5 & 94.5\\
  + PMP (Ours) & 21.4 {\color{blue} (+4.3)}  & 12.6 {\color{blue} (+3.6)} & 32.6 {\color{blue} (+7.1)}  & 14.7 {\color{blue} (+2.3)} & 49.5 {\color{blue} (+8.5)}  & 33.5 {\color{blue} (+3.9)} & 70.1 {\color{blue} (+3.8)}  & 57.3 {\color{blue} (+1.6)} & 97.4 {\color{blue} (+1.9)}  & 95.8 {\color{blue} (+1.3)} \\
  \midrule
  FC-CLIP (ConvNeXt-Large)~\cite{yu2023convolutions} & 21.5 & 14.8 & 29.6 & 18.2 & 45.7 & 34.1 & 69.2 & 58.4 & 96.1 &  95.4\\
  + PMP (Ours) & 25.5 {\color{blue} (+4.0)}  & {16.2} {\color{blue} (+1.4)} & 33.7 {\color{blue} (+4.1)}  & {19.1} {\color{blue} (+0.9)} & 52.6 {\color{blue} (+6.9)}  & {35.3} {\color{blue} (+1.2)} &73.8 {\color{blue} (+4.6)}  &  {59.9} {\color{blue} (+1.5)} & 97.9 {\color{blue} (+1.8)}  & {96.0} {\color{blue} (+0.6)}\\
  \bottomrule
  \end{tabular}
  }\label{table:two-stage}
\end{table*}
In order to analyze the situation that first-stage masks do not contain the masks that correspond to the text prompts (i.e, the ground-truth target masks), we provided and presented the result in Table~\ref{table:two-stage}. We conducted the ablations using two backbones: OVseg and FC-CLIP. We can observe that our model on top of either backbone achieves much  more performance gain in terms of the mIOU of ground truth in the first stage compared to the second stage (final result). To be clarify, the reason why the recall mIOU in the first stage is higher then second stage is because we calculate the recall for each ground truth against all of the unclassified proposals, which does not account for the error after association and classfication in second stage. This supports the claim that our model is leaning to generate more precise and accurate mask proposals in the first stage. Hence, if we can develop an improved matching algorithm in second stage, the performance can be further improved.

\paragraph{Backbones.}
\begin{table*}[t!]
  \centering
  \captionsetup{aboveskip=0pt}\captionsetup{belowskip=0pt}
  \caption{Ablation studies (The mIOU ($\%$) is utilized).}\label{table:abl_reb}
  \resizebox{0.75\linewidth}{!}
  {
  \begin{tabular}{l|ccccc}
  \toprule
  Method & ADE-847 & PC-459 & ADE-150 & PC-59 & VOC\\
  \midrule
  Default: OVSeg~\cite{liang2023open} (Swin-B) + PMP & {12.6} & {14.7} & {33.5} & {57.3} & {95.8} \\
  $\lambda=0.65$, $L=3$, $\lambda_{ce}=5.0, \lambda_{dice}=5.0$ &&&&\\
  \midrule
   OVSeg~\cite{liang2023open}(Swin-S)+ PMP  &{11.4} & {13.2} & {27.7} & {53.5} & {92.2} \\
   OVSeg~\cite{liang2023open}(Swin-L)+ PMP  &{13.4} & {15.6} & {34.7} & {58.1} & {96.1} \\
   OVSeg~\cite{liang2023open}(R101c) + PMP  & {9.1} & {12.5} & {25.5} & {52.4} & {91.9} \\
   OVSeg~\cite{liang2023open}(R101)+ PMP  & {8.6} & {11.9} & {24.9} & {51.8} & {91.2} \\
   OVSeg~\cite{liang2023open}(R50)+ PMP  &{8.1} & {11.7} & {24.5} & {51.3} & {91.0} \\


  \midrule
   OVSeg~\cite{liang2023open}($\lambda=0.5$) + PMP  & {12.5} & {14.7} & {33.5} & {57.4} & {95.6} \\
   OVSeg~\cite{liang2023open}($\lambda=0.6$) + PMP  & {12.6} & {14.7} & {33.5} & {57.3} & {95.7} \\
   OVSeg~\cite{liang2023open}($\lambda=0.7$) + PMP  & {12.6} & {14.6} & {33.4} & {57.3} & {95.8} \\

  \midrule
   OVSeg~\cite{liang2023open}($L=1$) + PMP  & {11.0} & {13.2} & {32.1} & {55.1} & {95.0} \\
   OVSeg~\cite{liang2023open}($L=2$) + PMP  & {12.1} & {14.5} & {33.4} & {57.0} & {95.5} \\
   OVSeg~\cite{liang2023open}($L=5$) + PMP  & {13.0} & {14.9} & {33.9} & {57.7} & {96.0} \\
  \midrule
   OVSeg~\cite{liang2023open}($\lambda_{ce}=7.0, \lambda_{dice}=3.0$) + PMP  & {12.7} & {14.7} & {33.4} & {57.3} & {95.8} \\
   OVSeg~\cite{liang2023open}($\lambda_{ce}=6.0, \lambda_{dice}=4.0$) + PMP  & {12.6} & {14.7} & {33.5} & {57.3} & {95.8} \\
   OVSeg~\cite{liang2023open}($\lambda_{ce}=4.0, \lambda_{dice}=6.0$) + PMP  & {12.6} & {14.7} & {33.5} & {57.3} & {95.6} \\
  
  \bottomrule
  \end{tabular}
  }
\end{table*}

To further analyze the importance of the backbone choice for the feature encoder, we ablate the backbones with Swin Transformer small (Swin-S), Swin Transformer large (Swin-L), ResNet-101 with 3x3 convolution (R101c), ResNet-101 (R101), ResNet-50 (R50) on top of OVSeg~\cite{liang2023open} in the second big row of Table~\ref{table:abl_reb}.

\paragraph{Hyperparameters}

We provide more ablations on some of the hyperparameters in the third, fourth, fifth rows. 
For the balancing factor $\lambda$ in Eq.~\ref{eq:3}, we set the default as $\lambda = 0.65$ following ODISE~\cite{xu2023open} for simplicity. For reference, OVSeg~\cite{liang2023open} set $\lambda = 0.7$ for A-150 and A-847, and set $\lambda = 0.6$ for PAS20, PC-59 and PC-459. We still provide ablation on this in the third row of Table~\ref{table:abl_reb}. For $L$ in transformer decoder, we used the default setting $L=3$ from the backbone Mask2Former~\cite{cheng2022masked} while provide more ablation studies on it in the fourth row of Table~\ref{table:abl_reb}. Similarly, following the original backbone of Mask2Former~\cite{cheng2022masked}, the loss weights are set as $\lambda_{ce}=5.0$ and $\lambda_{dice}=5.0$ equally while more ablations can be obtained in the fifth row of Table~\ref{table:abl_reb}.

For the temperature coefficient {$\tau$} in Eq.~\ref{eq:2}, it is the learnable parameter following CLIP~\cite{radford2021learning} thus it is directly optimized during training as a log-parameterized multiplicative scalar to avoid turning as a hyper-parameter. 

\subsection{More qualitative results}

\begin{figure*}[t!]
  \centering
  \includegraphics[width=\linewidth]{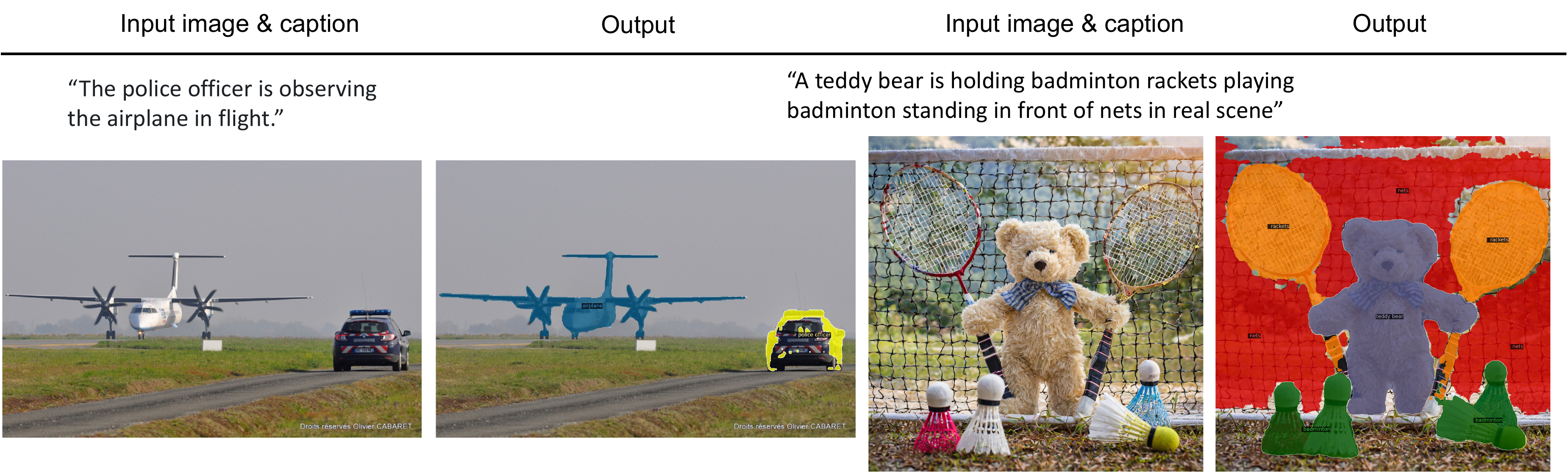}
  \captionsetup{aboveskip=0pt}\captionsetup{belowskip=0pt}
  \caption{The example of evaluation on a given caption. The left sample is chosen from the COCO test set while the right sample is the generated from the GenAI platform (Adobe Firefly).}
  \label{fig:cap}
\end{figure*}

\paragraph{\textbf{Captions as input.}}
To support that our model is able to segment given the single caption (a sentence), we present examples of open-vocabulary segmentation in Figure~\ref{fig:cap}. This also demonstracts that our model can be used for data for recognition and the data from generative AI.

\begin{figure*}[t!]
  \centering
  \includegraphics[width=\linewidth]{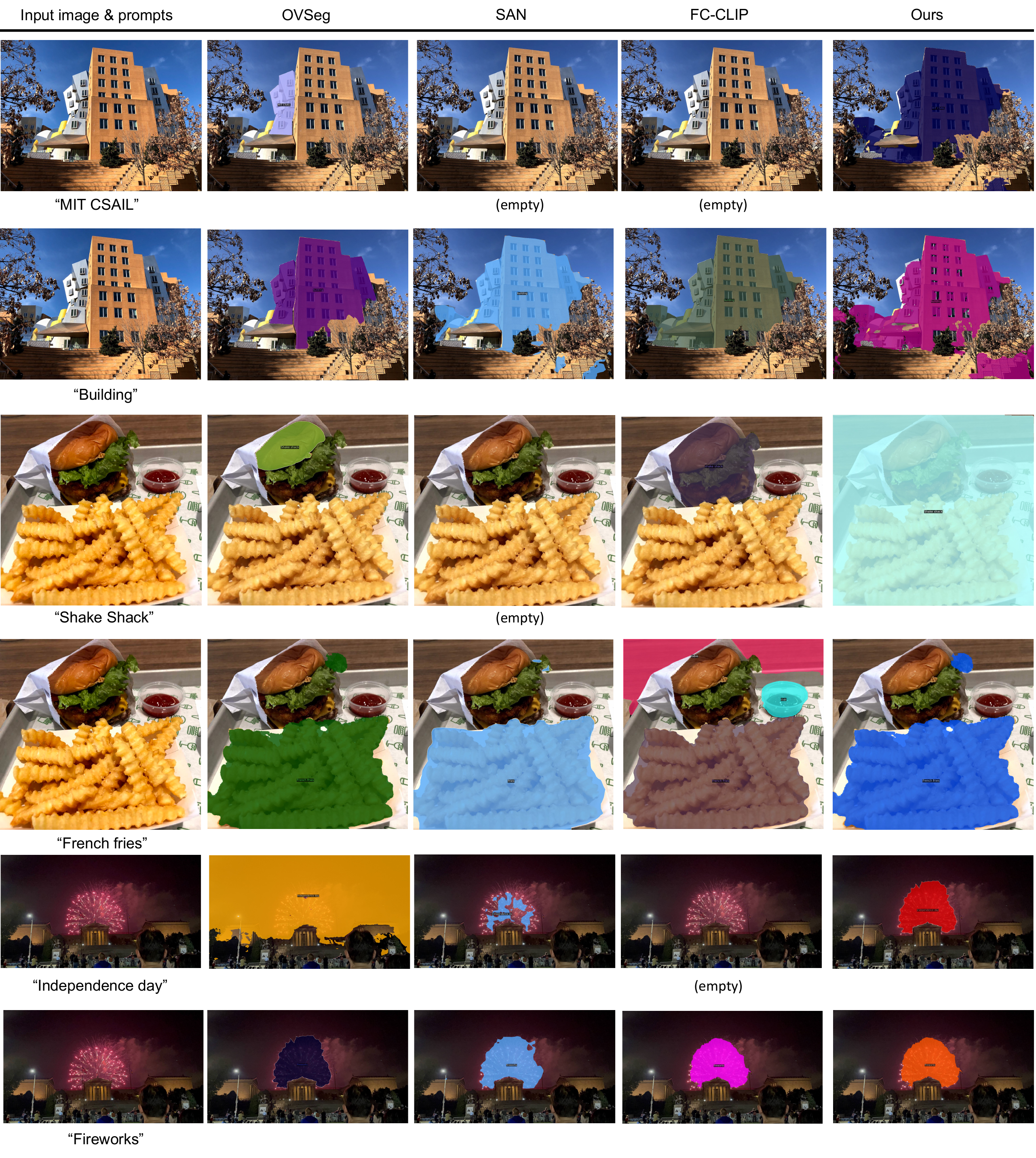}
  \caption{More qualitative results of open-vocabulary segmentation on our taken seven example real images. We compare our models further with OVSeg~\cite{liang2023open}, SAN~\cite{xu2023side}, and FC-CLIP~\cite{yu2023convolutions}. 
  }
  \label{fig:qual_sup}
\end{figure*}

\paragraph{\textbf{More comparisons.}}
To further support the claim that our Prompt-guided Mask Proposal (PMP) is able to handle abstract queries, we presented more results in Figure~\ref{fig:qual_sup} and compared with OVSeg~\cite{liang2023open}, SAN~\cite{xu2023side}, and FC-CLIP~\cite{yu2023convolutions}. The output produced by our pipeline is built on top of OVSeg since OVSeg is more generalized to difficult prompts qualitatively. Each of the inputs contains one image and one prompt where we ablate the difficulty of the input prompts given two different prompts: difficult and easy prompts. We can observe, among all of the compared models, the outputs produced by our PMP are able to capture the area correctly even when a difficult word such as ``MIT CSAIL'', ``Shake Shack'', or ``Independence Day" is given. On the other hand, most of the current methods perform satisfactorily on the easy prompts, which shows that their models are highly trained to capture easy words using their original pipeline. This result supports that our proposed PMP opens a new chapter for true open-vocabulary segmentation in real-world applications.

\begin{figure*}[t!]
  \centering
  \includegraphics[width=\linewidth]{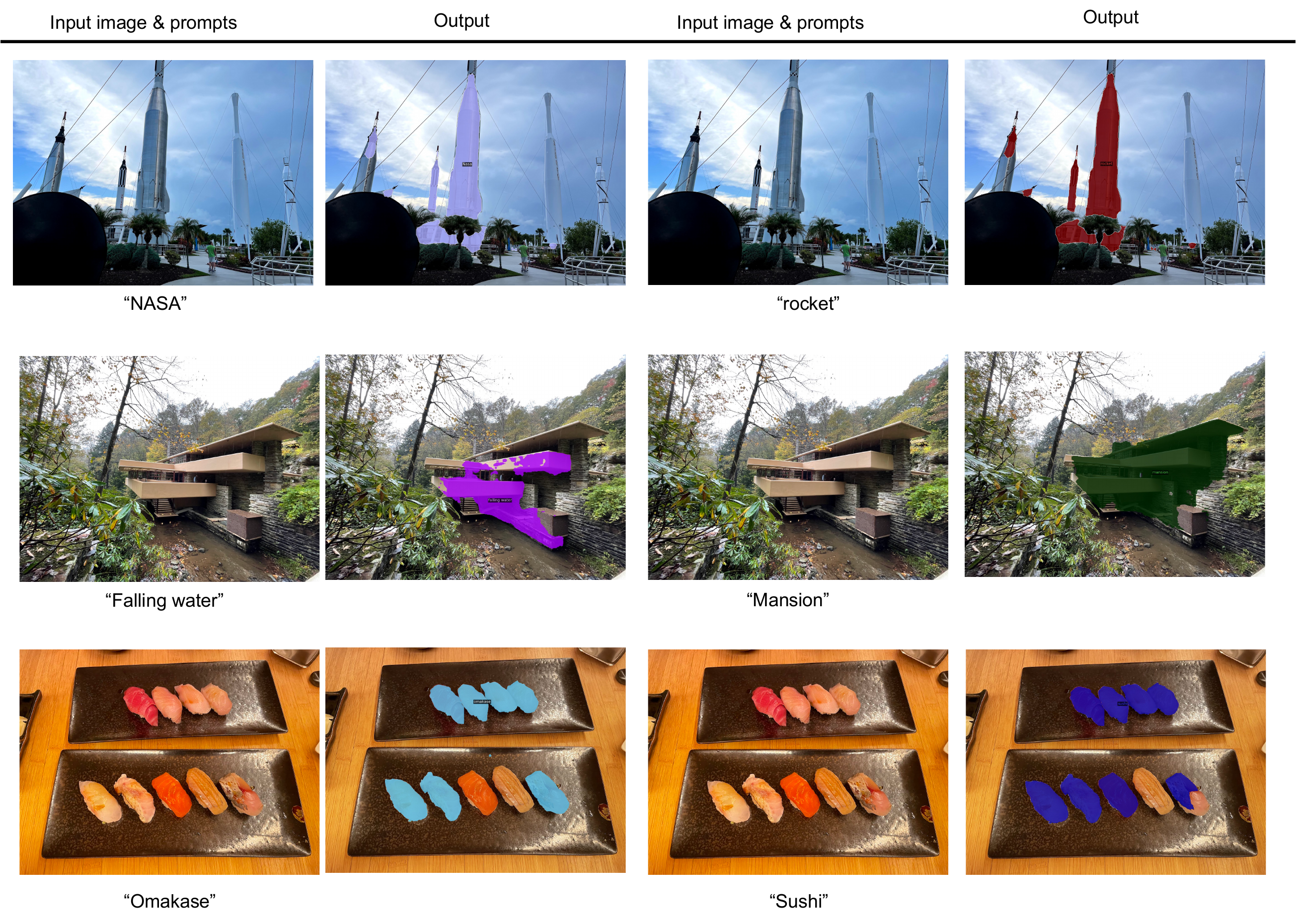}
  \caption{Failure cases of open-vocabulary segmentation on our taken seven example real images.
  }
  \label{fig:fail_sup}
\end{figure*}

\paragraph{\textbf{Failure cases.}}
Even though our PMP is capable of recognizing the area of abstract prompts, the quality of segmentation maps can still be improved which can not be reflected in the current five benchmarks. Thus, we presented several examples of the failure cases using our pipeline plus OVSeg~\cite{liang2023open} in Figure~\ref{fig:fail_sup}, showing the cases when our segmentation maps are not ideal. For example, the even though our model is able to connect ``NASA'' with ``'rocket', it still can not capture all regions of the area of rockets.

\subsection{More implementation details}

Since our proposed prompt-guided mask proposals (PMP) are built on top of Mask2Former~\cite{cheng2022masked} as the backbone for proposal generation in the first stage, we now provide the implementation details of the modules in Mask2Former here in the stage one.

\paragraph{\textbf{Image encoder.}}
Our image encoder is adaptable to any backbone architecture, akin to MaskFormer and Mask2Former. In this study, we utilized either standard convolution-based ResNet~\cite{he2016deep} backbones (R50 and R101 with 50 and 101 layers, respectively) or the recently introduced Transformer-based Swin-Transformer~\cite{liu2021swin} backbones, depending on the settings for a fair comparison with prior works. Further details can be found in \cite{cheng2021per,cheng2022masked}.

\paragraph{\textbf{Pixel decoder.}}
Similar to Mask2Former and MaskFormer, our pixel decoder is compatible with any existing pixel decoder module. This implies that it can be implemented using any semantic segmentation decoder (e.g., \cite{chen2018encoder,cheng2020panoptic}). 
The Transformer module attends to all image features, gathering global information to generate class predictions. This design reduces the necessity for a per-pixel module for extensive context aggregation. MaskFormer introduces a lightweight pixel decoder based on the widely used FPN~\cite{lin2017feature} architecture. In Mask2Former, the more advanced multi-scale deformable attention Transformer (MSDeformAttn)~\cite{zhu2020deformable} is used as the default pixel decoder, demonstrating superior results across various segmentation tasks.

\paragraph{\textbf{Transformer decoder.}}
We utilized the Transformer decoder with $L=3$ (i.e., 9 layers in total) and $100$ queries by default. An auxiliary loss is applied to every intermediate Transformer decoder layer and to the learnable query features before the Transformer decoder.

\paragraph{\textbf{Loss weights.}}
In line with \cite{cheng2022masked}, we employed binary cross-entropy loss and the dice loss~\cite{milletari2016v} for our mask loss: $\mathcal{L}_{mask} = \lambda_{ce} \mathcal{L}_{ce} + \lambda_{dice} \mathcal{L}_{dice}$. 
Note that the loss weights need to be set differently according to the backbone approach to be combined with our pipeline.

\subsection{Model Efficiency}

For 100 text tokens, the PMP pipeline has roughly 1.03 s instead of 30s on top of OVSeg~\cite{liang2023open}. \textit{The statistics of inference time can also be validated in their paper~\cite{liang2023open}}, which supports our correction of the inference report. The details can be summarized as follows:
\begin{itemize}
    \item OVseg (Swin-B) + PMP (ours) 1.03s: token extraction (0.22s) + first stage (0.21s) + second stage (0.6s).
    \item OVseg (Swin-B) 1.02s: token extraction (0.22s) + first stage (0.2s) + second stage (0.6s).
\end{itemize}
We can observe that our PMP only brings 0.1s in the first stage and bring obvious performance gain. We also observe the same phenomenon that the PMP does not bring obvious latency when combining with other existing works.

\subsection{Social Impact and limitation}
Our work on open-vocabulary segmentation has a significant social impact by vastly enhancing the ability of systems to recognize and interact with an unlimited range of categories, surpassing the limitations of existing models restricted to a predefined set of classes. Traditional segmentation approaches struggle to identify objects or concepts not included in their fixed vocabulary, limiting their ability to handle the dynamic and diverse nature of real-world scenarios. In contrast, our method empowers systems to understand any object or category described in natural language, enabling them to respond to open-ended prompts without being constrained by a finite list of labels. This flexibility makes technology more adaptable and accessible, allowing users to interact in a more natural way without the burden of knowing pre-defined categories to generate the mask proposal with our pipeline. 

However, it's important to acknowledge a limitation: while our approach generates masks that accurately reflect the objects described in text prompts, the precision of these masks can be limited. This makes the method excellent for understanding language-driven descriptions and general category recognition, but it may not be as suitable for tasks requiring fine-grained perception and precise delineation of object boundaries. As such, our method is more aligned with applications where broad understanding of language is prioritized over pixel-perfect accuracy, highlighting its strengths in adaptability and accessibility rather than in scenarios demanding high-precision segmentation.

\end{document}